\documentclass[sigconf,nonacm]{acmart}
\renewcommand\footnotetextcopyrightpermission[1]{}

\usepackage{cleveref}
\usepackage{algorithm}
\usepackage{algorithmic}
\usepackage[most]{tcolorbox}
\usepackage{soul} 
\usepackage{mdframed}
\usepackage{stfloats}
\usepackage{cuted}
\usepackage{capt-of} 

\usepackage{float}       
\usepackage{booktabs}    
\usepackage{caption}    
\usepackage{multirow}
\usepackage{bm}
\usepackage{booktabs}
\usepackage{threeparttable}

\newcommand{\mypm}[1]{{\scriptsize $\pm$ #1}}

\AtBeginDocument{%
  }

\begin{document}

\title{Modeling Subjective Urban Perception with Human Gaze}

\author{Lin Che}
\email{linche@ethz.ch}

\affiliation{%
  \institution{ETH Zurich}
  \country{Switzerland}
}

\author{Xi Wang}
\email{xi.wang@inf.ethz.ch}

\affiliation{%
  \institution{ETH Zurich}
  \country{Switzerland}
}

\author{Marc Pollefeys}
\email{marc.pollefeys@inf.ethz.ch}

\affiliation{%
  \institution{ETH Zurich}
  \country{Switzerland}
}

\author{Konrad Schindler}
\email{schindler@ethz.ch}

\affiliation{%
  \institution{ETH Zurich}
  \country{Switzerland}
}

\author{Martin Raubal}
\email{mraubal@ethz.ch}

\affiliation{%
  \institution{ETH Zurich}
  \country{Switzerland}
}

\author{Peter Kiefer}
\email{pekiefer@ethz.ch}

\affiliation{%
  \institution{ETH Zurich}
  \country{Switzerland}
}

\renewcommand{\shortauthors}{Che et al.}

\begin{abstract}
Urban perception describes how people subjectively evaluate urban environments, shaping how cities are experienced and understood. Existing computational approaches primarily model urban perception directly from street view images, but largely ignore the human perceptual process through which such judgments are formed. In this paper, we introduce \textit{Place Pulse-Gaze}, an urban perception dataset that augments street view images with synchronized eye-tracking recordings and individual perception labels. Based on this dataset, we propose a \textit{Gaze-Guided Urban Perception Framework} to study how gaze behavior contributes to the modeling of subjective urban perception. The framework systematically investigates three complementary settings: gaze-only modeling, gaze fusion with explicit semantic scene representations, and gaze fusion with implicit richer visual representations. Experiments show that gaze alone already carries useful predictive signals for subjective urban perception, and that integrating gaze with scene representations further improves prediction under both semantic and richer visual representations. Overall, our findings highlight the importance of incorporating human perceptual processes into urban scene understanding and open a direction for gaze-guided multimodal urban computing. The dataset and code will be available at \url{https://github.com/lin102/Place-Pulse-Gaze}.
\end{abstract}

\maketitle

\section{Introduction}
Urban perception describes how individuals subjectively evaluate and interpret urban environments, forming impressions \cite{salesses2013collaborative, ito2024understanding}. Human urban perceptions play a critical role in shaping urban experience, influencing residential choices, public health outcomes, economic activities, and policy-making \cite{nasar1990evaluative, kelling1982broken, cohen2000broken, ross2001neighborhood, dijksterhuis2001perception}. The proliferation of street view imagery, together with advances in computer vision, has enabled computational modeling of urban perception at an unprecedented scale. The Place Pulse 1.0 and 2.0 datasets from the MIT Media Lab \cite{salesses2013collaborative, dubey2016deep} marked a turning point in computational urban perception research. By collecting large-scale crowd-sourced annotations on geotagged street view images, the dataset enabled the development of computer vision models capable of predicting multiple perceptual attributes (safety, wealth, liveliness, beauty, boredom, and depression) directly from street view content. This image-based paradigm has also enabled researchers to systematically examine how urban appearance relates to social outcomes such as public health, crime rates and mobility patterns \cite{park2020pedestrian, fu2018streetnet, li2023integrating}. 

However, most existing approaches implicitly treat the perceptual impression of an urban image on humans as an objective property of the image itself, modeling it as a direct mapping from pixels to perception labels \cite{porzi2015predicting, min2019multi, moreno2021quantifying}. This image-centric formulation overlooks the fundamentally human-centered nature of perception: urban impressions do not arise solely from environmental content, but also from how individuals allocate visual attention and cognitively interpret cues within the scene. Recent studies suggest that the predominantly image-based formulation of urban perception is limited, as perceptual judgments systematically vary across individual demographic attributes and personality traits, underscoring the inherently human-centered and subjective nature of urban perception \cite{quintana2024my, quintana2025global}. 

Individual subjectivity is also reflected in the perceptual process itself. Human visual behavior, particularly patterns of gaze and attention allocation, has long been regarded as a window into higher-level cognitive states \cite{yarbus1967eye, henderson2013predicting}. Eye tracking provides a direct and quantitative means to capture how individuals explore visual environments, revealing their attentional strategies and interpretative focus \cite{cavanagh2011visual}. Prior work has leveraged eye-tracking signals for a range of cognition-related tasks, including Alzheimer’s disease detection \cite{sriram2023classification}, egocentric activity recognition \cite{ozdel2024gaze}, scene understanding \cite{henderson2011eye}, and aesthetic preference prediction \cite{pappas2020quickly}. The rapid advancement of eye-tracking technology and the emergence of wearable devices such as smart glasses further suggest strong potential for large-scale, real-world deployment of attention-aware modeling \cite{novak2024eye}. However, despite the inherently human-centered nature of urban perception, existing approaches rarely account for individuals’ subjective visual exploration processes. 

To bridge this gap, we make the following contributions:

\begin{itemize}
    \item We introduce \textit{Place Pulse-Gaze}, an urban perception dataset built upon a curated subset of Place Pulse 2.0, containing over 10k image-gaze pairs with individual gaze recordings and subjective perception labels.
    \item We propose a unified \textit{Gaze-Guided Urban Perception Framework} that models subjective urban perception either from gaze dynamics alone or by jointly integrating gaze with scene representations.
    \item Through extensive experiments, we show that gaze alone already carries useful predictive signals for subjective urban perception, and that integrating gaze with visual scene representations consistently improves performance over image-only baselines across different representation settings.
\end{itemize}

\section{Related Work}
\subsection{Urban Perception}
Urban perception has long been recognized as a critical factor shaping human behavior and well-being in cities \cite{dijksterhuis2001perception}. Understanding how people perceive urban environments is essential for improving urban experience and informing planning decisions \cite{lynch1964image, ito2024understanding}. Traditional urban studies often relied on field surveys and manual environmental assessments, which are costly, time-consuming, and difficult to scale \cite{gobster2004human, dadvand2016green}. The increasing availability of large-scale street view imagery has transformed this landscape, enabling scalable visual analysis of urban environments. Street views have been widely adopted in urban studies for tasks such as land-use classification, public health analysis, tourist recommendations, and accessibility evaluation \cite{hou2024global, che2025unsupervised, kang2020review, kubota2025omni, wang2024assessing}. 

A major milestone in computational urban perception research was the introduction of the Place Pulse 1.0 dataset \cite{salesses2013collaborative, naik2014streetscore}, which collected crowd-sourced perceptual labels (safety, class, and uniqueness) on geotagged street view images and demonstrated correlations between perceived safety and crime statistics. Building upon this data-collecting paradigm, Place Pulse 2.0 (PP2) \cite{dubey2016deep} substantially expanded the scale of data collection, covering 110,998 images from 56 cities worldwide and annotating them across six perceptual attributes: safety, wealth, liveliness, beauty, boredom, and depression. In addition to the dataset and benchmark, \citet{dubey2016deep} also introduced an end-to-end convolutional neural network that directly predicts perceptual attributes from visual content. Subsequent work by \citet{yao2019human} proposed a human–machine adversarial scoring framework to efficiently assess local urban perception, combining Fully Convolutional Networks with Random Forest models to improve prediction robustness. \citet{wang2022measuring} combined street-view imagery, deep learning, and space syntax theory to assess street spatial quality at scale. Similarly, \citet{dai2021analyzing} leveraged semantic segmentation of street view images combined with multivariate linear regression to examine the correlation between urban visual space and residents’ psychological perceptions, revealing that environmental features such as greenness and enclosure significantly influence subjective evaluations. Building upon street view-based safety perception models, \citet{ceccato2025makes} employed regression analysis and integrated income and crime statistics to study the relationship between street types and human safety perception. Despite methodological differences, these image-based approaches primarily rely on visual content and overlook individual-level subjective differences and the human-centered perceptual processes through which urban impressions are formed. In contrast, our work explicitly incorporates individual gaze behavior as an observable signal of the human perceptual process, enabling the modeling of subjective urban perception beyond image content alone.

\subsection{Eye-Tracking in Human Perception Modeling}
Recent evidence suggests that urban perception is not universal but rather varies systematically across individual profiles, including demographic attributes and personality traits \cite{quintana2025global}. This motivates moving beyond purely image-centric modeling: street view visual content alone may be insufficient to fully characterize human-centered subjective differences, and modeling the perceptual process itself can provide complementary signals for understanding how perceptions are formed and for improving prediction. A large body of work in vision and cognitive psychology has established eye movements and visual attention allocation as a behavioral proxy for high-level cognitive processes, reflecting how observers actively sample information during perception \cite{henderson2003human, rayner2009eye, krejtz2018eye}. Eye tracking has therefore been widely used to study and model subjective cognition-related tasks. For example, gaze has been shown to both correlate with and causally influence preference formation during decision-making \cite{shimojo2003gaze}, and fixation-based computational models have been proposed to explain binary choice behavior \cite{krajbich2010visual}. Beyond decision tasks, eye-movement patterns have been exploited to infer task and cognitive states \cite{henderson2013predicting}, and to recognize everyday activities using gaze-derived features such as saccades, fixations, and blinks \cite{bulling2010eye}. Eye tracking has also been applied to spatial cognition tasks \cite{montello2013functions} such as map reading and wayfinding \cite{kiefer2017eye}, further supporting its utility in capturing attention-driven perceptual strategies.

In the context of urban studies and the built environment, a growing set of works has started to incorporate eye tracking to analyze how people experience urban scenes. Prior studies report that attention to specific elements (e.g., construction-related objects or trash) is associated with stress and negative emotions \cite{tavakoli2025psycho}, and that certain eye-movement statistics (e.g., longer average saccade duration) correlate with higher satisfaction regarding pleasantness and perceived safety \cite{wang2025emotion}. \citet{crosby2019does} reported that during safety perception judgments, observers exhibit longer fixation durations on buildings, houses, and vehicles. Eye tracking has also been used to study landscape evaluation and preference in urban green spaces, where attention to trees and pedestrians is linked to more positive assessments \cite{li2020evaluation}. More recently, \citet{kang2026decoding} explored explainability for street view-based safety prediction by leveraging gaze heatmaps and comparing multiple explainable AI methods, finding that XGradCAM and EigenCAM most closely align with human safety perceptual patterns.

Despite these advances, most existing urban related studies use eye tracking primarily for correlational analysis or post-hoc attention visualization, rather than directly incorporating gaze dynamics into subjective urban perception modeling. The work most closely related to ours is \citet{yang2024urban}, which extracts fixation-based Area of Interest (AOI) statistics (e.g., total fixation duration, number of fixations, time to first fixation, and first fixation duration) from semantic segmentation outputs and combines them with image semantics for urban perception prediction using a random forest model. While promising, this approach still relies on aggregated statistical features and largely ignores the sequential dynamics inherent in eye-movement behavior. Moreover, it operates exclusively on semantic representations derived from image segmentation, disregarding the possibility that lower-level visual features might offer richer perceptual cues. In contrast, our work introduces a framework that jointly models gaze sequences and street view images for urban perception prediction. Further, we release the Place Pulse-Gaze dataset to support attention-aware and individualized urban perception research.

\begin{figure*}[t]
\centering
\includegraphics[width=0.95\textwidth]{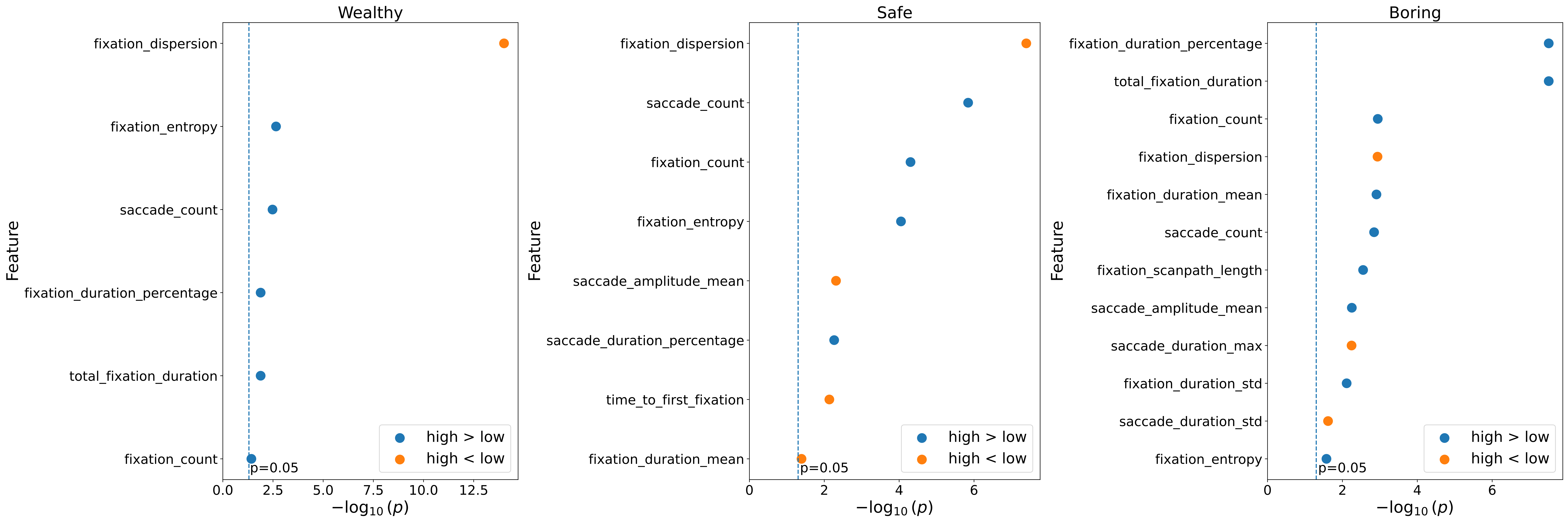} 
\caption{Significant gaze-only features under one-way ANOVA across perception levels (Low/Neutral/High). Features with $p<0.05$ are shown; the dashed line marks the $p=0.05$ threshold in $-\log_{10}(p)$. Blue indicates High$>$Low and orange indicates High$<$Low.}
\label{fig_gaze_only_anova}
\end{figure*}

\begin{figure*}[t]
\centering
\includegraphics[width=0.95\textwidth]{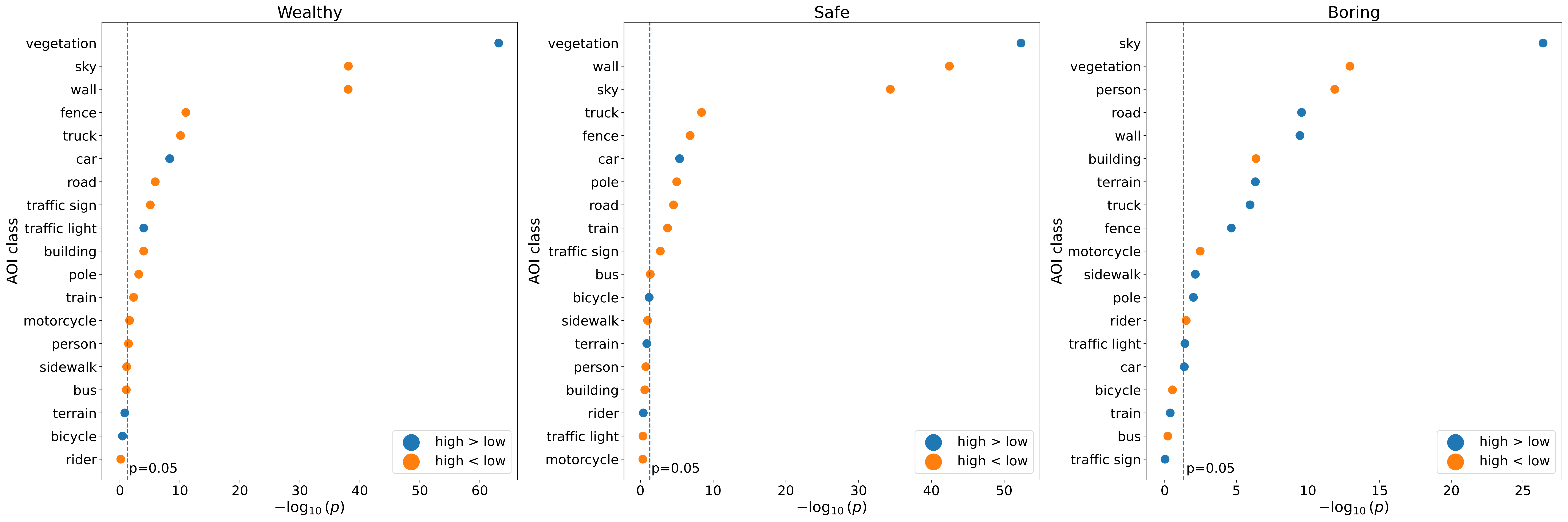} 
\caption{Significant AOI fixation features under one-way ANOVA across perception levels (Low/Neutral/High). The dashed line marks the $p=0.05$ threshold in $-\log_{10}(p)$. Blue indicates High$>$Low and orange indicates High$<$Low in mean fixation time.}
\label{fig_gaze_aoi_anova}
\end{figure*}

\section{Place Pulse-Gaze Dataset}
We construct \textit{Place Pulse-Gaze}, a gaze-augmented urban perception dataset built upon a curated subset of \textit{Place Pulse 2.0}. It enriches street view images with jointly collected eye-tracking recordings and corresponding perception labels, enabling research on attention-aware, individual-level urban perception. The study was approved by the ETH Zurich Ethics Commission. 

\subsection{Image Selection and Processing}
Due to the time and effort needed to collect eye-tracking data, it was infeasible to include all 110,998 images of the large-scale Place Pulse 2.0 dataset in our eye tracking study. We adopt a quota-based sampling strategy to choose a balanced yet manageable subset of images. Quotas are based on the score distribution for three perception dimensions. It has been shown that several perception dimensions exhibit strong correlations \cite{dubey2016deep}. Therefore, in order to reduce redundancy and lower the workload of participants during the eye-tracking experiment, we focus on the following three relatively less correlated dimensions: Wealth, Safe, and Boredom. We first remove samples with missing perception scores. Then, for each perception dimension, the score distribution is divided into ten equal-width bins between the minimum and maximum values. We then randomly sample 80 images from each bin to ensure coverage across the full perception spectrum. Applying this procedure across the three perception dimensions results in a total of 2,248 street view images. 

To improve visual clarity for participants and enable more accurate mapping of gaze coordinates onto the images, we upsample the original low-resolution images (400×300 pixels) using a super-resolution model \cite{wang2021real} to 1600×1100 pixels.

\subsection{Eye-Tracking Study}
We recruited 96 participants aged between 18 and 55 years for the eye-tracking study. Eye movements were recorded using a Tobii Pro Spectrum eye tracker (Tobii AB, Sweden) at a sampling rate of 600\,Hz, following the manufacturer’s recommended experimental setup and recording procedures~\cite{tobii_spectrum}. A standard calibration procedure was performed prior to the experiment for each participant. During each trial, participants were instructed to view a street view image displayed on a 24-inch monitor for 7 seconds. After the viewing phase, they rated the image along three perception dimensions: wealthy, safe, and boring. Following common practice in psychophysics studies, ratings were collected using a 5-point Likert scale~\cite{valtchanov2015cognitive}, where 3 indicates a neutral perception, values above 3 indicate positive evaluations, and values below 3 indicate negative evaluations.

To reduce visual fatigue, participants were required to take a mandatory break of at least one minute after every 10 trials. Each participant completed 125 trials, resulting in an average session duration of approximately one hour. Each image was viewed and rated by five different participants, each providing both a perception rating and a corresponding gaze recording. At the end of the experiment, participants completed a post-study questionnaire collecting demographic information (age and gender), personality traits measured using the Ten-Item Personality Inventory (TIPI), and residential background, including the countries and continents, and the types of environments they had lived in or preferred to live in (urban, suburban, or rural). 

We removed invalid recordings and samples with excessively low valid gaze ratios. After filtering, the final dataset contains 10,223 valid image-gaze pairs from 96 participants over 2,248 street view images, which are used for subsequent analysis and modeling.

\subsection{Inter-Rater Perception Variability}
\label{subsec:inter_rater_variability}
We examine the variability of perceptual judgments across participants viewing the same image. To characterize this variability, we discretize the original 5-point Likert ratings into three levels: Low (rating < 3), Neutral (rating = 3), and High (rating > 3), consistent with our ternary prediction setup. We then compute \textit{Krippendorff’s $\alpha$} to quantify inter-rater agreement and analyze the distribution of \textit{Mean Pairwise Distance} (MPD) between raters for each image \cite{krippendorff2018content}.
Across the three perception dimensions, the \textit{Boring}  dimension exhibits the lowest inter-rater agreement ($\alpha$ = 0.170, 95\% CI [0.150, 0.189]), followed by \textit{Safe} ($\alpha$ = 0.370, 95\% CI [0.348, 0.392]), while \textit{Wealthy} shows the highest agreement ($\alpha$ = 0.504, 95\% CI [0.482, 0.524]). The MPD distributions, as shown in \Cref{fig:label_consistency} (Appendix), further illustrate the variability distribution. The Boring dimension exhibits higher variability, likely due to its subjective and interpretation-dependent nature, while Wealthy tend to be anchored in more consistently perceived visual cues. These results reveal notable individual differences in urban perception, with varying levels of variability across perceptual dimensions, suggesting the importance of modeling perception at the individual level.

\section{Gaze-Perception Analysis}

To better understand how eye-movement behavior relates to subjective urban perception, we use one-way ANOVA as an exploratory analysis to identify gaze and AOI features that vary across perception levels, rather than to establish causal attentional determinants. Specifically, we analyze (i) gaze-only features and (ii) semantic AOI-based attention patterns. The former characterizes \textbf{how} participants visually explore a scene based solely on gaze dynamics, while the latter characterizes \textbf{what} they attend to by quantifying fixation allocation over different semantic elements of the urban environment.

\subsection{Gaze and Image Processing}

\subsubsection{Gaze Event Detection}
\label{subsec:gaze_event_detection}
Following common practice, raw gaze recordings (600 Hz, 7 s) are converted into fixation and saccade events using the I-DT fixation detection algorithm \cite{ salvucci2000identifying, duchowski2017eye}. These events provide behaviorally meaningful units for subsequent sequence modeling and analysis.

\subsubsection{Semantic Image Segmentation}
To obtain semantic scene representations for gaze-scene analysis, we performed pixel-wise semantic segmentation on all street view images. Specifically, we adopted Mask2Former~\cite{cheng2022masked} pretrained on the Cityscapes dataset~\cite{cordts2016cityscapes} to generate dense semantic label maps covering the 19 standard urban semantic categories defined in Cityscapes (Table~\ref{tab:semantic_categories}, Appendix). These semantic masks allow us to associate visual attention with scene objects in the subsequent AOI-based analysis.

\subsection{Gaze-Only Feature Analysis}
\label{subsec:gaze_only_feature_analysis}
To characterize \textbf{how} participants visually explore street view scenes under different perceptual levels, we extract 21 standard gaze features  (Table~\ref{tab:gaze_features}, Appendix), capturing global eye-movement dynamics across fixations, saccades, and scanpaths \cite{duchowski2017eye, mahanama2022eye, selim2024review}. We conduct univariate one-way ANOVA to test whether each feature differs significantly across perception levels (Low/Neutral/High) for each perception dimension. For features with significant ANOVA results, we further conduct post-hoc Tukey HSD tests to examine pairwise differences across perception levels.

Figure~\ref{fig_gaze_only_anova} summarizes features with $p<0.05$. We find that multiple gaze features vary systematically across perception levels, such as fixation dispersion and fixation count, indicating that perceptual differences are accompanied by distinct visual exploration patterns even without explicitly modeling image content. Detailed post-hoc Tukey HSD results are provided in Table~\ref{tab:gaze_tukey_summary} (Appendix). These observations provide behavioral evidence that motivates our subsequent modeling of gaze dynamics for urban perception prediction.

\subsection{Semantic AOI-based Attention Analysis}
\label{subsec:semantic_AOI_based_attention_analysis}
We next analyze \textbf{what} semantic elements participants attend to when forming urban perception judgments. Using the semantic segmentation masks, we treat each semantic category as a type of AOI and compute, for each image, the proportion of total fixation time allocated to each AOI. We then perform one-way ANOVA to test whether AOI fixation time allocation over the 19 classes differs across perception levels. The results are shown in Fig.~\ref{fig_gaze_aoi_anova}. For AOIs with significant ANOVA effects, we further conduct post-hoc Tukey HSD tests to examine pairwise differences between perception levels; the corresponding results are summarized in Table~\ref{tab:aoi_tukey_summary} (Appendix).

For \textit{Wealthy} and \textit{Safe}, the most significant semantic attention patterns are broadly consistent. In particular, higher fixation allocation to \textit{vegetation} is associated with higher perceived safety and wealth, whereas increased attention to \textit{wall}, \textit{sky}, \textit{truck}, and \textit{fence} is associated with lower perceived safety and wealth. In contrast, \textit{Boring} exhibits a distinct pattern: greater attention to \textit{sky}, \textit{road}, and \textit{wall} corresponds to higher perceived boredom, while attention to \textit{vegetation} and \textit{person} corresponds to lower perceived boredom.

Taken together, the gaze-only and AOI-based analyses suggest that subjective urban perception is accompanied by systematic differences in both \textbf{how} observers explore a scene and \textbf{what} semantic elements receive attention. These findings motivate our subsequent gaze-guided framework that jointly models gaze sequences and street view content for individualized urban perception prediction.

\begin{figure*}[t]
\centering
\includegraphics[width=0.98\textwidth]{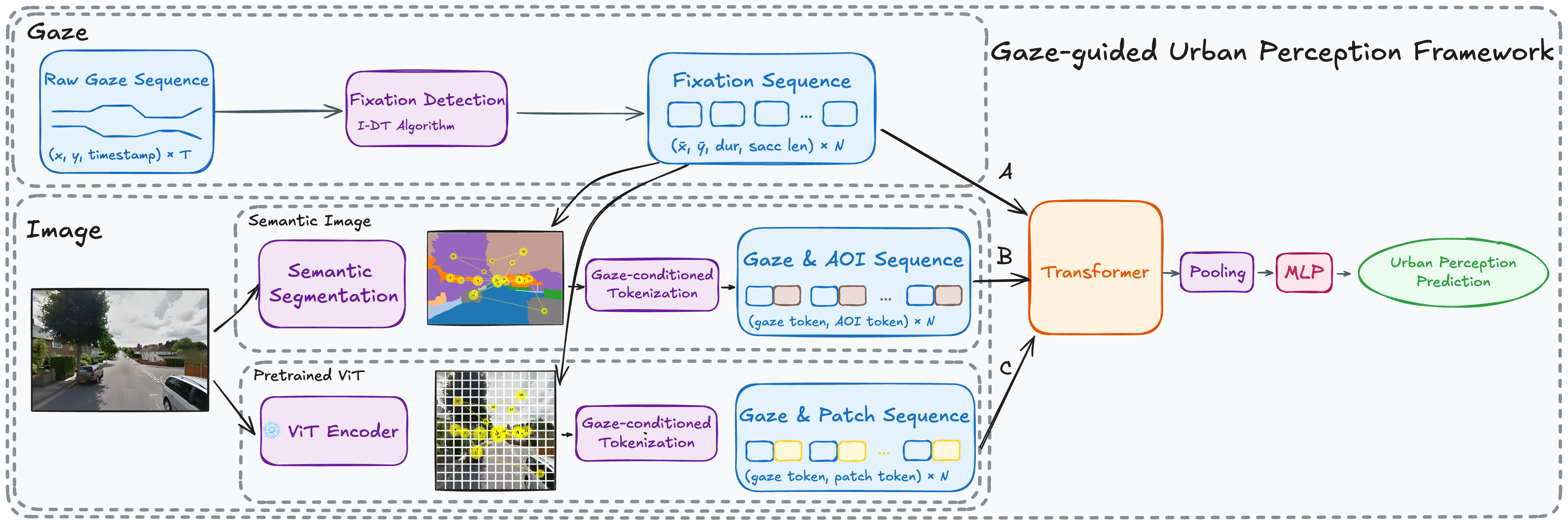} 
\caption{
Overview of the proposed \textbf{Gaze-guided Urban Perception Framework}. 
Raw gaze recordings are first segmented into fixation sequences using the I-DT algorithm. The fixation sequence is then used to construct token sequences for three modeling variants: 
(A) \textit{Gaze-only modeling}, which represents perception purely from gaze dynamics; 
(B) \textit{Gaze + Semantic AOI fusion}, where gaze tokens are paired with semantic scene tokens obtained from semantic segmentation; and 
(C) \textit{Gaze + ViT patch fusion}, where gaze tokens are paired with visual patch representations extracted by a pretrained ViT. 
All variants are formulated as sequence modeling problems and processed by a shared Transformer encoder followed by pooling and an MLP head for urban perception prediction.
}
\label{fig:framework}
\end{figure*}

\section{Method}

\subsection{Problem Formulation}
As shown in Sec.~\ref{subsec:gaze_only_feature_analysis} and \ref{subsec:semantic_AOI_based_attention_analysis}, subjective urban perception is associated with both \textbf{how} observers visually explore a scene and \textbf{what} semantic elements they attend to. This suggests that image-centric approaches relying solely on scene content may be insufficient to fully capture subject-specific perception differences. Motivated by this observation, we formulate the subjective urban perception task as follows.

Given a street-view image $i$, a subject $s$, and a perception dimension $k \in \{\textit{Wealth}, \textit{Safety}, \textit{Boredom}\}$, our goal is to predict the subject-specific perception level $y_{i,s}^{(k)} \in \{\textit{Low}, \textit{Neutral}, \textit{High}\}$. We view the observed ordinal rating as a discretization latent perception score $z_{i,s}^{(k)}$, and, for conceptual clarity, describe it as

\begin{equation}
z_{i,s}^{(k)} = f\!\left(\mu_i^{(k)}, \Delta_{i,s}^{(k)}\right) + \epsilon_{i,s}^{(k)}
\label{eq:decomposition_eq}
\end{equation}

\noindent where $\mu_i^{(k)}$ denotes a consensus scene-driven component associated with visual cues that tend to be interpreted similarly across observers, $\Delta_{i,s}^{(k)}$ represents subject-specific deviations from this consensus induced by individual attention allocation and perceptual interpretation, and $\epsilon_{i,s}^{(k)}$ captures residual noise, including random rating fluctuations and subject-specific bias. Here, $f(\cdot,\cdot)$ captures how the consensus component and the subject-specific component jointly determine the perceived score, without assuming a specific functional form. This formulation is intended only as a conceptual abstraction of factors contributing to subject-specific urban perception, rather than as an explicitly estimated generative model.

Under this formulation, visual scene representations primarily capture the consensus scene-driven component $\mu_i^{(k)}$. Gaze behavior, provides complementary information about the perceptual process, potentially reflecting both common viewing patterns across observers and subject-specific variations. Motivated by this view, we propose a unified gaze-guided urban perception framework that models subjective urban perception either from gaze dynamics alone or jointly with scene representations, enabling both gaze-only and multimodal prediction settings.

\subsection{Gaze-Guided Urban Perception Framework}

Based on the formulation above, we propose a subject-specific, \textit{Gaze-guided Urban Perception Framework}, as illustrated in Fig.~\ref{fig:framework}. Unlike image-centric formulations, our framework explicitly incorporates the perceptual process through gaze behavior, enabling the modeling of subjective urban perception at the individual level. Given a subject's raw gaze recording while viewing a street view image, the framework predicts the perceived level of the target qualities. Raw gaze signals are first converted into fixation events, which serve as behaviorally meaningful units for all subsequent modeling variants.

We structure the framework around three complementary modeling questions: 
\textbf{(1) whether gaze dynamics alone already carries some signal about (subject-specific) urban perception; 
\textbf{(2)} whether gaze improves perceptual modeling when combined with explicit semantic scene representations; and 
\textbf{(3)} whether gaze improves perceptual modeling when combined with richer, low-level visual representations. }

To answer these questions, we instantiate the framework under three variants, each corresponding to one setting. 
\textbf{(A) Gaze-only modeling} uses fixation-based gaze tokens without any scene information to evaluate the prediction of gaze alone. 
\textbf{(B) Gaze + Semantic AOI fusion} combines gaze tokens with explicit semantic scene tokens derived from image segmentation, enabling gaze-scene fusion grounded in semantic object categories. 
\textbf{(C) Gaze + ViT patch fusion} combines gaze tokens with pretrained visual patch representations extracted from a Vision Transformer (ViT), to examine whether gaze also anhances perception modeling with a comprehensive representation of the visual content. 
Although the scene representations differ, all three variants share the same gaze tokenization pipeline and Transformer-based sequence modeling backbone. From the perspective of Eq.~\ref{eq:decomposition_eq}, the gaze-only variant leverages perceptual cues contained in gaze behavior, including subject-specific variations associated with $\Delta$, whereas the multimodal variants combine gaze with scene representations to better capture richer consensus scene-driven component $\mu$ and subject-specific deviations.

\subsection{Gaze-Only Modeling}
\label{subsec:gaze_only_modeling}
We first investigate whether gaze dynamics alone carry any noticeable signal that predicts the subject-specific urban perception; corresponding to Variant A in Fig.~\ref{fig:framework}, respectively to $\Delta$ in Eq.~\ref{eq:decomposition_eq}. Following the preprocessing pipeline described in Sec.~\ref{subsec:gaze_event_detection}, raw gaze recordings are segmented into a sequence of fixation events.

Following prior gaze sequence modeling work \cite{lohr2022eye, chen2021predicting}, each fixation event is represented as a gaze token using its mean spatial location $(\bar{x}, \bar{y})$, fixation duration, and the length of the subsequent saccade. These token-level descriptors capture key spatial and temporal characteristics of visual exploration. The resulting feature vectors are projected into a 128-dimensional latent embedding space and fed into a Transformer encoder to model dependencies across gaze events. The encoded sequence is then aggregated by a mean pooling layer and passed to an MLP head to predict the subject-specific urban perception level.

\subsection{Multimodal Gaze--Image Fusion}
Under the decomposition in Eq.~\ref{eq:decomposition_eq}, multimodal fusion aims to combine scene-driven cues related to $\mu$ with gaze-derived subject-specific cues associated with $\Delta$. While the gaze-only variant evaluates whether eye-movement dynamics alone carry predictive signals for urban perception, we further investigate whether gaze behavior provides complementary subject-specific cues beyond scene appearance by fusing gaze tokens with image-derived scene representations.

In the multimodal variants, we preserve an event-level sequential formulation by pairing each gaze token with a scene token derived from the corresponding image. In this way, the resulting multimodal sequence jointly captures \textit{how} the observer explores the scene and \textit{what} scene information is being attended to. We instantiate this fusion strategy with two types of scene representations: \textbf{(B) semantic AOI tokens} derived from explicit semantic segmentation, and \textbf{(C) visual patch tokens} extracted from a pretrained Vision Transformer. Both variants share the same multimodal sequence modeling backbone and differ only in the form of scene token used for fusion, as illustrated in Fig.~\ref{fig:framework}.

\subsubsection{\textbf{Gaze + Semantic AOI Fusion}}
\label{subsec:gaze_semantic_aoi}
As illustrated by Variant B in Fig.~\ref{fig:framework}, we first obtain semantic segmentation maps over the 19 AOI categories using the same image processing pipeline described in Sec.~\ref{subsec:semantic_AOI_based_attention_analysis}. This variant explicitly grounds gaze behavior in semantic scene elements, enabling interpretable gaze-scene fusion. We then perform \textit{gaze-conditioned semantic tokenization}: for each fixation event, its mean spatial location $(\bar{x}, \bar{y})$ is used to assign the fixation to a semantic AOI category, yielding a sequence of AOI labels aligned with the fixation sequence.

Each AOI label is embedded into a 128-dimensional semantic token, which is concatenated with the corresponding 128-dimensional gaze token to form a 256-dimensional multimodal token. The resulting multimodal token sequence is then processed by the same Transformer backbone as in the gaze-only variant.

\subsubsection{\textbf{Gaze + Pretrained ViT Patch Fusion}}

As illustrated by Variant C in Fig.~\ref{fig:framework}, we investigate whether gaze provides complementary perceptual cues when combined with strong pretrained visual representations for urban perception prediction. We extract patch-level visual embeddings from a frozen Vision Transformer pretrained on ImageNet-21k \cite{wu2020visual}. 

We then perform \textit{gaze-conditioned patch tokenization}, where each fixation is assigned to the corresponding image patch according to its spatial location $(\bar{x}, \bar{y})$, producing a sequence of patch tokens aligned with the fixation sequence. Fusion with gaze tokens follows the same multimodal token construction described in Sec.~\ref{subsec:gaze_semantic_aoi}.

\section{Experiments}

\subsection{Experimental Setup}

We perform an image-level split of the dataset into training, validation, and test sets with a ratio of 70\% / 15\% / 15\%. To avoid leakage across splits, all image--gaze pairs associated with the same image are assigned to the same split. Each perception dimension is modeled independently to enable dimension-specific analysis of gaze and image contributions, and to avoid potential interference across tasks.

We report both Macro-F1 and Accuracy, and use Macro-F1 as the primary evaluation metric to account for class imbalance in the three-level perception classification task. All models are trained using the standard cross-entropy loss and the AdamW optimizer with a peak learning rate of $1\times10^{-4}$ and a batch size of 128. We train for 30 epochs using a cosine learning rate schedule with 1.5 epochs of linear warmup. Model selection is based on the best validation Macro-F1. All results are reported as the mean and standard deviation over five runs with different random seeds.

\begin{table*}[t]
\centering
\small
\setlength{\tabcolsep}{5pt}
\begin{tabular}{lcccccc}
\toprule
\multirow{2}{*}{\textbf{Token Representation}} 
& \multicolumn{2}{c}{\textbf{Wealthy}} 
& \multicolumn{2}{c}{\textbf{Safe}} 
& \multicolumn{2}{c}{\textbf{Boring}} \\
\cmidrule(lr){2-3} \cmidrule(lr){4-5} \cmidrule(lr){6-7}
& \textbf{Macro-F1  $\uparrow$} & \textbf{ACC $\uparrow$} 
& \textbf{Macro-F1 $\uparrow$} & \textbf{ACC $\uparrow$} 
& \textbf{Macro-F1 $\uparrow$} & \textbf{ACC $\uparrow$} \\
\midrule
$(\bar{x}, \bar{y})$
& 36.8 \mypm{1.2} & 37.7 \mypm{0.8} 
& 36.1 \mypm{0.5} & 39.5 \mypm{0.6} 
& 36.0 \mypm{0.3} & 38.6 \mypm{1.0} \\

$(\bar{x}, \bar{y})$ + duration 
& 39.0 \mypm{1.4} & 39.6 \mypm{1.1} 
& 36.9 \mypm{0.6} & 39.1 \mypm{0.9} 
& 37.4 \mypm{1.3} & 38.8 \mypm{1.1} \\

\textbf{\bm{$(\bar{x}, \bar{y})$} + duration + saccade length}
& \textbf{40.1} \mypm{0.6} & \textbf{40.3} \mypm{0.6} 
& \textbf{37.4} \mypm{0.6} & \textbf{40.0} \mypm{0.8} 
& \textbf{38.4} \mypm{0.7} & \textbf{39.8} \mypm{0.4} \\
\bottomrule
\end{tabular}

\caption{Effect of fixation event token representation in the gaze-only model. All values are reported in percentage (\%). Best scores are in \textbf{bold}.}
\label{tab:gaze_only_input_design}
\end{table*}

\begin{table*}[t]
\centering
\small
\setlength{\tabcolsep}{6pt}

\begin{tabular}{lcccccc}
\toprule
& \multicolumn{2}{c}{\textbf{Wealthy}} 
& \multicolumn{2}{c}{\textbf{Safe}} 
& \multicolumn{2}{c}{\textbf{Boring}} \\
\cmidrule(lr){2-3} \cmidrule(lr){4-5} \cmidrule(lr){6-7}
\textbf{Model}
& \textbf{Macro-F1} $\uparrow$ & \textbf{ACC} $\uparrow$
& \textbf{Macro-F1} $\uparrow$ & \textbf{ACC} $\uparrow$
& \textbf{Macro-F1} $\uparrow$ & \textbf{ACC} $\uparrow$ \\
\midrule

Fixation Heatmap 
& 35.2 \mypm{0.9} & 36.0 \mypm{1.2}
& 35.0 \mypm{1.5} & 37.4 \mypm{1.6}
& 33.8 \mypm{1.6} & 39.1 \mypm{1.4} \\

XGBoost 
& 38.1 \mypm{0.9} & 38.4 \mypm{0.9}
& 37.0 \mypm{1.0} & 39.2 \mypm{1.0}
& 37.5 \mypm{0.5} & 37.6 \mypm{0.5} \\

\textbf{Gaze-Only Transformer }
& \textbf{40.1} \mypm{0.6} & \textbf{40.3} \mypm{0.6}
& \textbf{37.4} \mypm{0.6} & \textbf{40.0} \mypm{0.8}
& \textbf{38.4} \mypm{0.7} & \textbf{39.8} \mypm{0.4} \\
\bottomrule
\end{tabular}

\caption{Comparison of gaze-only modeling methods. All values are reported in percentage (\%). Best scores are in \textbf{bold}.}
\label{tab:gaze_only_baselines}
\end{table*}

\subsection{Gaze-Only Modeling Results}
\label{subsec:gaze_only_res}
We first investigate how different \textit{gaze token representations} affect fixation-event sequence modeling. Specifically, while keeping the 2-layer 4-head Transformer backbone fixed, we compare three token configurations: fixation mean location $(\bar{x}, \bar{y})$ alone, mean location with fixation duration, and mean location with both fixation duration and subsequent saccade length. 

As shown in Table~\ref{tab:gaze_only_input_design}, progressively adding duration and saccade length consistently improves performance across all three perception dimensions. This indicates that urban perception is related not only to \textit{where} observers look, but also to the temporal dynamics of \textit{how} they inspect the scene. These findings support the use of event-level behavior sequential gaze modeling for subject-specific urban perception prediction.

Based on the above results, we adopt the full fixation-event representation $(\bar{x}, \bar{y}, \text{duration}, \text{saccade length})$ in the final gaze-only model. We compare it with two baselines: (i) a fixation heatmap baseline that aggregates fixation locations into a Gaussian spatial map and feeds it to a CNN-based image classifier, and (ii) an XGBoost baseline \cite{chen2016xgboost} using the 21 hand-crafted gaze features listed in Table~\ref{tab:gaze_features}.

Table~\ref{tab:gaze_only_baselines} shows that the fixation heatmap baseline performs only marginally above chance level (33.3\% Macro-F1 for three-way classification), indicating that spatial fixation locations alone (\textit{where}) are insufficient for subjective urban perception prediction. The XGBoost baseline achieves improved performance, which may be attributed to its ability to capture coarse temporal statistical cues related to \textit{how} observers inspect the scene. Our proposed \textit{Gaze-Only Transformer} further improves performance across all three perception dimensions. However, the overall performance remains relatively low, i.e., gaze alone provides some signal, but only a weak one. This is expected, as gaze only reflects the attentional allocation process but not the visual content itself. Nevertheless, the consistent gains indicate that modeling event-level \textit{where}+\textit{how} information and sequential dependencies better captures the perceptual cues in gaze data compared to static spatial representations or aggregated statistical statistics.

\begin{table*}[t]
\centering
\small
\setlength{\tabcolsep}{6pt}

\begin{tabular}{lcccccc}
\toprule
& \multicolumn{2}{c}{\textbf{Wealthy}} 
& \multicolumn{2}{c}{\textbf{Safe}} 
& \multicolumn{2}{c}{\textbf{Boring}} \\
\cmidrule(lr){2-3} \cmidrule(lr){4-5} \cmidrule(lr){6-7}
\textbf{Model}
& \textbf{Macro-F1} $\uparrow$ & \textbf{ACC} $\uparrow$
& \textbf{Macro-F1} $\uparrow$ & \textbf{ACC} $\uparrow$
& \textbf{Macro-F1} $\uparrow$ & \textbf{ACC} $\uparrow$ \\
\midrule
AOI-Only Model
& 40.5 \mypm{1.2} & 42.5 \mypm{0.8}
& 39.2 \mypm{0.3} & 42.1 \mypm{0.3}
& 39.9 \mypm{0.5} & \textbf{43.8} \mypm{0.6} \\

AOI Sequence Transformer
& 41.8 \mypm{0.2} & 43.0 \mypm{0.6}
& 39.6 \mypm{0.6} & 43.2 \mypm{0.6}
& 40.3 \mypm{0.4} & 41.9 \mypm{0.5} \\

\textbf{Gaze + AOI Transformer}
& \textbf{46.1} \mypm{0.4} & \textbf{46.3} \mypm{0.5}
& \textbf{42.2} \mypm{0.4} & \textbf{45.0} \mypm{0.3}
& \textbf{41.1} \mypm{0.4} & 42.6 \mypm{0.6} \\
\midrule

w/o Gaze
& 41.9 \mypm{0.9} & 42.9 \mypm{0.8}
& 39.4 \mypm{0.4} & 43.4 \mypm{0.3}
& 39.6 \mypm{1.0} & 41.2 \mypm{0.7} \\

Shuffled AOI Alignment
& 37.4 \mypm{1.3} & 37.7 \mypm{1.3}
& 36.6 \mypm{1.0} & 39.5 \mypm{0.9}
& 37.6 \mypm{0.6} & 38.9 \mypm{0.7} \\
\bottomrule
\end{tabular}

\caption{Comparison of Semantic AOI-only and gaze-semantic AOI fusion. All values are reported in percentage (\%).  Best scores are in \textbf{bold}. The upper block compares AOI-only and gaze-semantic fusion models, while the lower block reports ablations on the proposed Gaze + AOI Transformer.}
\label{tab:gaze_aoi_results}
\end{table*}

\begin{table*}[t]
\centering
\small
\setlength{\tabcolsep}{6pt}

\begin{tabular}{lcccccc}
\toprule
& \multicolumn{2}{c}{\textbf{Wealthy}} 
& \multicolumn{2}{c}{\textbf{Safe}} 
& \multicolumn{2}{c}{\textbf{Boring}} \\
\cmidrule(lr){2-3} \cmidrule(lr){4-5} \cmidrule(lr){6-7}
\textbf{Model}
& \textbf{Macro-F1} $\uparrow$ & \textbf{ACC} $\uparrow$
& \textbf{Macro-F1} $\uparrow$ & \textbf{ACC} $\uparrow$
& \textbf{Macro-F1} $\uparrow$ & \textbf{ACC} $\uparrow$ \\
\midrule

Image-Only ViT
& 56.3 \mypm{0.6} & 56.2 \mypm{0.4}
& 49.1 \mypm{0.8} & 50.7 \mypm{1.0}
& 44.8 \mypm{0.4} & 46.2 \mypm{0.5} \\

Gaze-weighted Patch Pooling
& 54.8 \mypm{1.6} & 54.8 \mypm{1.2}
& 48.8 \mypm{1.0} & 50.9 \mypm{1.2}
& 44.0 \mypm{0.4} & 45.0 \mypm{0.7} \\

Patch Sequence Transformer
& 56.8 \mypm{1.2} & 56.9 \mypm{0.8}
& 49.9 \mypm{1.3} & 51.9 \mypm{0.7}
& 43.3 \mypm{0.8} & 44.9 \mypm{0.7} \\

\textbf{Gaze + Patch Transformer}
& \textbf{58.1} \mypm{0.5} & \textbf{58.0} \mypm{0.4}
& \textbf{50.5} \mypm{0.4} & \textbf{52.1} \mypm{0.8}
& \textbf{45.5} \mypm{0.6} & \textbf{46.6} \mypm{0.7} \\
\midrule

w/o Gaze
& 56.7 \mypm{1.1} & 56.7 \mypm{0.9}
& 50.1 \mypm{1.0} & 51.7 \mypm{1.0}
& 44.5 \mypm{0.8} & 46.3 \mypm{1.3} \\

Shuffled Patch Alignment
& 56.9 \mypm{0.7} & 56.7 \mypm{0.7}
& 49.9 \mypm{1.0} & 50.9 \mypm{1.2}
& 44.9 \mypm{1.0} & 46.4 \mypm{1.5} \\
\bottomrule
\end{tabular}

\caption{Comparison of image-only ViT and gaze-fusion models based on pretrained ViT patch representations.
The upper block compares image-only, sequence-only, and gaze-weighted fusion baselines, while the lower block reports ablations on the proposed Gaze + Patch Transformer. 
All values are reported in percentage (\%). Best scores are in \textbf{bold}.}
\label{tab:gaze_patch_results}

\end{table*}

\subsection{Gaze + Semantic AOI Fusion Results}

We next evaluate whether gaze improves perception prediction when combined with semantic scene representations derived from image segmentation (Table~\ref{tab:gaze_aoi_results}).

We first compare two semantic-only baselines. The \textit{AOI-only Image} model, following \citet{yang2024urban}, represents each image using a semantic composition vector. The \textit{AOI Sequence Transformer} constructs a sequence of semantic AOI tokens ordered by gaze fixations, but does not incorporate gaze tokens. The proposed \textit{Gaze + AOI Transformer} further concatenates gaze tokens with the corresponding semantic AOI tokens, enabling joint modeling of gaze behavior and viewed scene semantics. Models use the same 1-layer 4-head Transformer backbone.

As shown in Table~\ref{tab:gaze_aoi_results}, the AOI-only Image baseline performs slightly better than the gaze-only Transformer in Table~\ref{tab:gaze_only_baselines}, suggesting that explicit semantic scene structure provides strong cues for urban perception prediction. Compared with the AOI-only baseline, the AOI Sequence Transformer yields a small improvement. When gaze features are further incorporated, the proposed \textbf{Gaze + AOI Transformer} achieves the best performance on all three perception attributes. This result shows that gaze provides complementary perceptual cues beyond semantic scene content alone. In particular, the fusion model jointly captures \textit{where} observers allocate attention, \textit{how} they inspect the scene through temporal gaze dynamics, and \textit{what} semantic elements are being attended to. 

To further disentangle the role of gaze in the proposed fusion model, we first evaluate a capacity-matched \textit{w/o Gaze} ablation, where gaze-token inputs are replaced with zeros while keeping the multimodal architecture unchanged. Its lower performance in Table~\ref{tab:gaze_aoi_results} shows that the gain of the full model is not merely due to increased model capacity, but depends on informative gaze signals. We also examine whether this gain further depends on meaningful gaze--scene correspondence by randomly shuffling the AOI tokens assigned to fixation events (\textit{Shuffled AOI Alignment}). This destroys the spatial correspondence between gaze and scene semantics while preserving the overall token distribution and model architecture. As shown in Table~\ref{tab:gaze_aoi_results}, performance drops substantially, confirming that the improvement is not simply due to the presence of additional semantic tokens, but depends critically on the correct alignment between gaze behavior and scene semantics.

These findings are also consistent with our conceptual formulation in Eq.~\ref{eq:decomposition_eq}. The semantic AOI representation captures the scene-driven shared component $\mu$, whereas gaze contains complementary subject-specific cues related to $\Delta$. Their fusion therefore provides a more complete representation of subjective urban perception.

\subsection{Gaze + Pretrained ViT Fusion Results}

We further investigate whether gaze still provides complementary predictive cues when the image modality is included in a richer, less abstracted representation, namely the output of a pretrained, frozen ViT encoder. As shown in Table~\ref{tab:gaze_patch_results}, the \textit{Image-Only ViT} baseline substantially outperforms the AOI-only baseline in Table~\ref{tab:gaze_aoi_results}. This suggests that large-scale pretrained visual representations provide a much stronger approximation of the shared scene-driven component $\mu$ in Eq.~\ref{eq:decomposition_eq}, leading to markedly better perception prediction from image content alone.

\begin{figure*}[t]
\centering
\includegraphics[width=0.95\textwidth]{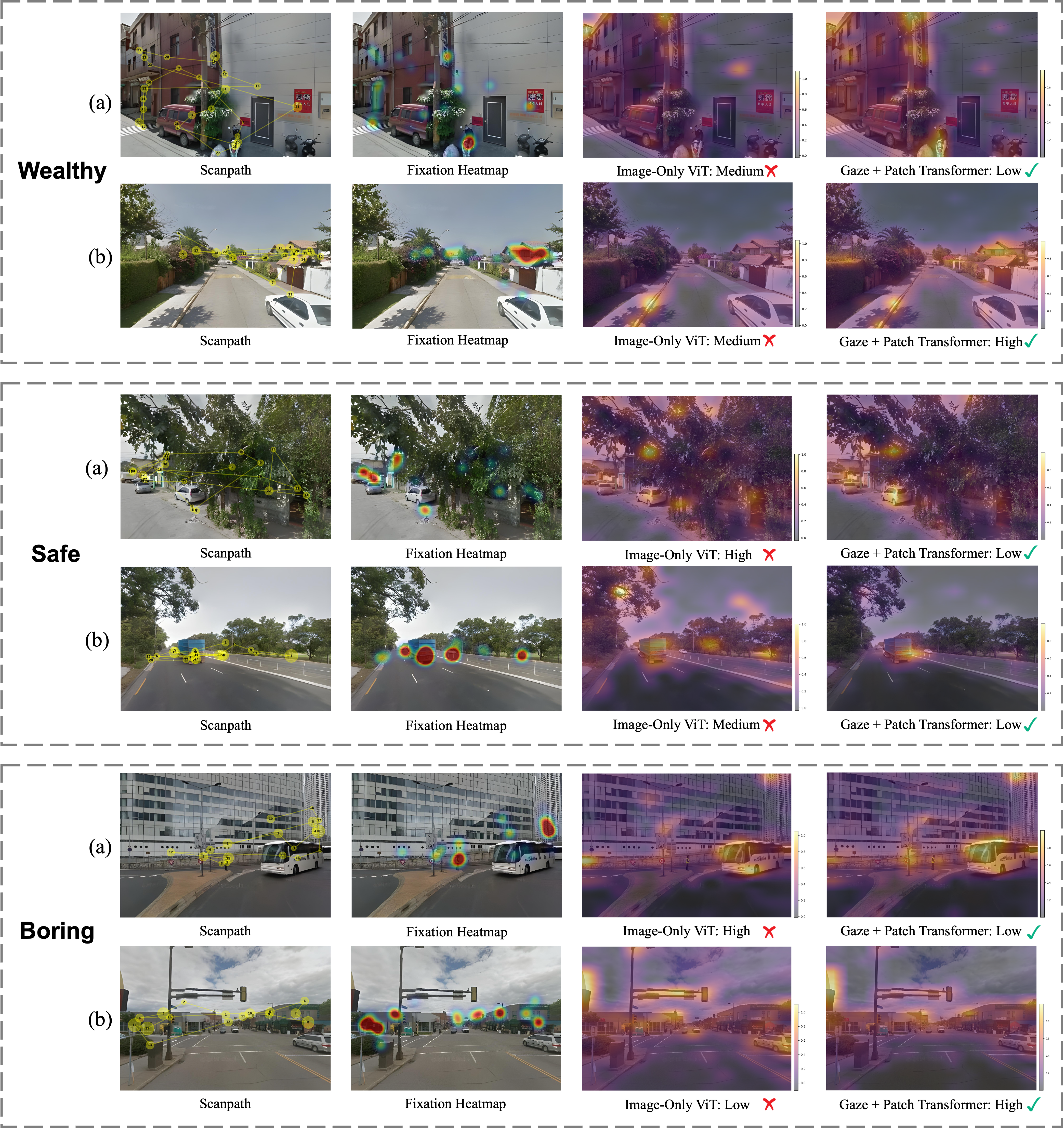} 
\caption{
Qualitative attribution comparison of the \textit{Image-Only ViT baseline} and the proposed \textit{Gaze + Patch Transformer} on an image misclassified by the former but correctly classified by the latter. We show the Scanpath, Fixation Heatmap, and patch-level attributions, computed using Layer Integrated Gradients on the predicted logit.}
\label{fig:IG_Vis}
\end{figure*}

We also compare two additional baselines. The first is Gaze-weighted Patch Pooling, a fusion strategy that uses a fixation heat map as a soft spatial prior to reweight patch features before pooling \cite{li2021eye}. The second is the Patch Sequence Transformer, which constructs a sequence of viewed patch tokens according to gaze order but does not incorporate explicit gaze token features. As shown in Table~\ref{tab:gaze_patch_results}, gaze-weighted Patch pooling performs worse than the Image-Only ViT baseline. This suggests that it is inadequate to simplify urban perception to importance weighting of the image cues. In addition, the Patch Sequence Transformer shows no clear gains over the Image-Only ViT baseline, indicating that patch viewing order alone provides insufficient subject-specific information once the scene-driven visual representation is strong enough.

Despite this stronger visual backbone, our proposed \textit{Gaze + Patch Transformer} still achieves the best performance across all three dimensions. This result indicates that gaze contributes complementary subject-specific cues beyond what is captured by image features, consistent with the role of the residual perceptual component $\Delta$ in Eq.~\ref{eq:decomposition_eq}. Compared with the gains observed in gaze-semantic AOI fusion, however, the improvements here are smaller. A likely explanation is that the pretrained ViT encoder already captures a larger fraction of the shared perceptual cues from the scene, leaving less room for additional gains from gaze. Moreover, the optimization is inherently asymmetric: the visual branch starts from a pretrained representation learned from internet-scale data, whereas the gaze branch has to be trained from scratch using a comparatively small set of gaze recordings, so that the joint training might tend to over-rely on the image information.

We further include the same ablations for the proposed \textit{Gaze + Patch Transformer}. The \textit{w/o Gaze} ablation shows consistently lower performance, confirming that the gain is not solely due to increased model capacity but relies on informative gaze signals. Shuffling the gaze-patch correspondence (\textit{Shuffled Patch Alignment}) also leads to a performance drop, indicating that alignment remains beneficial. The degradation is less pronounced than in the shuffled AOI alignment experiment. This likely reflects the richer contextual nature of pretrained ViT patch embeddings: through self-attention, each patch token already encodes some information from other image regions, making the model more tolerant to local alignment perturbations than in the explicit AOI-based setting.

\subsection{Qualitative Analysis of Gaze-Guided Patch Attributions}

We further conduct a qualitative analysis to understand how the proposed \textit{Gaze + Patch Transformer} improves prediction by leveraging gaze information on top of frozen ViT patch tokens. To this end, we employ Layer Integrated Gradients (LIG)~\cite{sundararajan2017axiomatic, kokhlikyan2020captum} to compute patch-level attribution scores for the predicted logits. Specifically, we attribute predictions to the initial patch embedding layer of the ViT backbone, i.e., the linear projection layer that maps image patches into patch embeddings. The resulting patch attributions are then projected back to the image and visualized as smoothed attribution maps.

We analyze cases where the \textit{Gaze + Patch Transformer} corrects mistakes made by the \textit{Image-Only ViT} baseline. Some examples are shown in Fig.~\ref{fig:IG_Vis}. Each row corresponds to one gaze recording. Across all three dimensions, the \textit{Image-Only ViT} tends to produce relatively diffuse attribution patterns, often spreading attention over many visually salient elements in the scene. In contrast, after incorporating gaze, the attribution maps become more concentrated on a smaller number of perceptually relevant regions. For example, in \textit{Safe (b)}, the image-only model distributes attribution across broad scene elements such as trees, sky, and vehicles, whereas the gaze-fusion model focuses more strongly on the truck region. Similarly, in \textit{Boring (b)}, the gaze-fusion model reduces emphasis on multiple traffic-related regions and shifts more attributes toward the building area, which may better support the final boring prediction.

In other cases, gaze also appears to redirect model attribution toward a different subset of scene elements. For instance, in \textit{Wealthy (a)}, attribution shifts toward the scooter region after incorporating gaze, while in \textit{Wealthy (b)}, attribution becomes more concentrated on the house. These examples do not by themselves establish causal perceptual determinants, but they illustrate that gaze guidance can alter which regions are prioritized by the model and may help it focus on relevant content among the many visual elements in a street view.

At the same time, the attribution maps are related to, but not identical to, the scanpath maps and fixation heat maps. This indicates that the \textit{Gaze + Patch Transformer} does not simply reweight or mask image patches according to gaze density. Instead, it appears to use gaze tokens together with their sequential dependencies to reshape patch-level evidence selection in a more complex and perception-relevant way. Finally, we note that the present qualitative analysis visualizes attributions only on the image branch. It does not directly quantify the contribution of gaze tokens themselves, which remains an interesting direction for future work.

\section{Limitations and Future Work}

While our study demonstrates the value of gaze for modeling subjective urban perception, several limitations remain. First, although \textit{Place Pulse 2.0} is one of the most widely used benchmarks for urban perception research, its images were collected in the early 2010s and may not fully reflect current urban conditions. In addition, the original image resolution is relatively low. Although we applied super-resolution to improve visual clarity for the eye-tracking experiment, this process may still introduce artifacts that could affect fine-grained viewing behavior in some cases. Furthermore, due to the distribution of the original dataset, the selected images remain geographically imbalanced, with much stronger coverage of Europe and North America than of Asia, Oceania and Africa.

Second, the high cost of collecting eye-tracking data limits the scale of the resulting dataset. Larger-scale gaze datasets would likely improve the robustness and generalizability of gaze-guided urban perception modeling. More broadly, our study is conducted under controlled laboratory conditions using static street view images. While this setting enables reliable measurement of gaze behavior, it does not fully capture the complexity of real-world urban perception in dynamic outdoor environments. We view this work as an initial step toward gaze-guided, subject-specific modeling of urban perception. Future work could continue in this direction and extend it towards larger-scale, in-the-wild settings. This includes also the integration of additional human physiological signals to better capture the perceptual and affective processes underlying urban experience.

\section{Conclusion}

We have studied how human gaze patterns can improve the modeling of subjective perception beyond image content alone, in the context of urban environments. To this end, we introduce the \textit{Place Pulse-Gaze}, an urban perception dataset that combines street view images with synchronized gaze recordings and individual perception labels, and propose a \textit{Gaze-Guided Urban Perception Framework}, supporting both gaze-only and multimodal perception modeling.

Our experiments show that gaze alone already carries some predictive signal, and that integrating gaze with scene representations further improves performance across both explicit semantic and pretrained visual representations. These findings highlight the importance of incorporating human perceptual processes into urban scene understanding and suggest promising directions for future gaze-guided multimodal modeling.

\bibliographystyle{ACM-Reference-Format}
\bibliography{sample-base}

\clearpage

\onecolumn 

\appendix
\section*{\Large Appendix}
\label{sec:appendix}

\section{Dataset and Analysis}

\subsection{Inter-rater Variability Distribution}
\begin{figure}[H]
    \centering
    \includegraphics[width=0.95\textwidth]{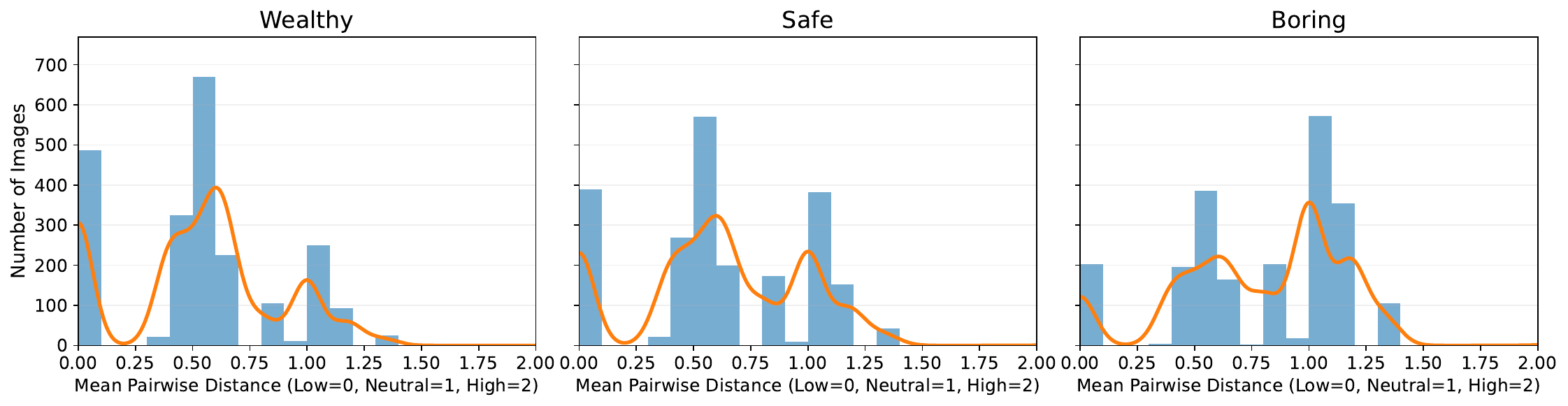}
    \caption{Distribution of Mean Pairwise Distance (MPD) between participants for the three perception dimensions. Ratings are discretized into Low (0), Neutral (1), and High (2). Larger MPD values indicate greater disagreement among participants.}
    \label{fig:label_consistency}
\end{figure}

Figure~\ref{fig:label_consistency} provides the full Distribution of Mean Pairwise Distance distributions for the three perception dimensions, complementing the discussion in Sec.~ \ref{subsec:inter_rater_variability}.

\subsection{Semantic Categories for AOI Analysis}
Table~\ref{tab:semantic_categories} lists the 19 semantic categories used in the Semantic AOI-based attention analysis in Sec.~ \ref{subsec:semantic_AOI_based_attention_analysis}.

\begin{table}[H]
    \centering
    \small 
    \begin{tabular}{rlp{1cm}rl}
        \toprule
        \textbf{ID} & \textbf{Label} & & \textbf{ID} & \textbf{Label} \\
        \midrule
        0 & road          & & 10 & sky           \\
        1 & sidewalk      & & 11 & person        \\
        2 & building      & & 12 & rider         \\
        3 & wall          & & 13 & car           \\
        4 & fence         & & 14 & truck         \\
        5 & pole          & & 15 & bus           \\
        6 & traffic light & & 16 & train         \\
        7 & traffic sign  & & 17 & motorcycle    \\
        8 & vegetation    & & 18 & bicycle       \\
        9 & terrain       & &    &               \\
        \bottomrule
    \end{tabular}
    \caption{Full list of the 19 semantic categories from the Mask2Former model.}
    \label{tab:semantic_categories}
\end{table}

\subsection{Gaze-Only Feature Definitions}
Table~\ref{tab:gaze_features} lists the 21 gaze-only features used in Sec.~ \ref{subsec:gaze_only_feature_analysis} and in the XGBoost baseline of Sec.~ \ref{subsec:gaze_only_res}.

\begin{table}[H]
    \centering
    \small
    \begin{tabular}{rlp{1.5cm}rl} %
        \toprule
        \textbf{No.} & \textbf{Gaze-Only Feature} & & \textbf{No.} & \textbf{Gaze-Only Feature} \\
        \midrule
        1  & fixation\_dispersion           & & 12 & total\_fixation\_duration      \\
        2  & saccade\_count                 & & 13 & fixation\_duration\_percentage \\
        3  & fixation\_count                & & 14 & fixation\_duration\_var        \\
        4  & fixation\_entropy              & & 15 & saccade\_duration\_max         \\
        5  & saccade\_amplitude\_mean       & & 16 & saccade\_duration\_std         \\
        6  & saccade\_duration\_percentage  & & 17 & saccade\_duration\_mean        \\
        7  & time\_to\_first\_fixation      & & 18 & saccade\_duration\_var         \\
        8  & fixation\_duration\_mean       & & 19 & saccade\_amplitude\_var        \\
        9  & fixation\_scanpath\_length     & & 20 & saccade\_amplitude\_std        \\
        10 & fixation\_duration\_max        & & 21 & saccade\_amplitude\_max        \\
        11 & fixation\_duration\_std        & &    &                                \\
        \bottomrule
    \end{tabular}
    \caption{List of Gaze-only features.}
    \label{tab:gaze_features}
\end{table}

\subsection{Post-hoc Tukey HSD Results}
Tables~\ref{tab:gaze_tukey_summary} and~\ref{tab:aoi_tukey_summary} summarize the pairwise post-hoc Tukey HSD comparisons for the significant ANOVA results reported in Secs.~\ref{subsec:gaze_only_feature_analysis} and ~\ref{subsec:semantic_AOI_based_attention_analysis}.

\begin{table*}[t]
\centering
\small
\setlength{\tabcolsep}{6pt}

\begin{tabular}{lccc ccc ccc}
\toprule
& \multicolumn{3}{c}{\textbf{Wealthy}} 
& \multicolumn{3}{c}{\textbf{Safe}} 
& \multicolumn{3}{c}{\textbf{Boring}} \\
\cmidrule(lr){2-4} \cmidrule(lr){5-7} \cmidrule(lr){8-10}
\textbf{Feature}
& \textbf{H--L} & \textbf{H--M} & \textbf{L--M}
& \textbf{H--L} & \textbf{H--M} & \textbf{L--M}
& \textbf{H--L} & \textbf{H--M} & \textbf{L--M} \\
\midrule

fixation\_count
& $\checkmark$ & \textit{n.s.} & \textit{n.s.}
& $\checkmark$ & $\checkmark$ & \textit{n.s.}
& \textit{n.s.} & $\checkmark$ & \textit{n.s.} \\

fixation\_dispersion
& $\checkmark$ & \textit{n.s.} & $\checkmark$
& $\checkmark$ & $\checkmark$ & $\checkmark$
& $\checkmark$ & \textit{n.s.} & \textit{n.s.} \\

fixation\_duration\_mean
& -- & -- & --
& \textit{n.s.} & \textit{n.s.} & \textit{n.s.}
& \textit{n.s.} & $\checkmark$ & $\checkmark$ \\

fixation\_duration\_percentage
& $\checkmark$ & \textit{n.s.} & \textit{n.s.}
& -- & -- & --
& \textit{n.s.} & $\checkmark$ & $\checkmark$ \\

fixation\_duration\_std
& -- & -- & --
& -- & -- & --
& \textit{n.s.} & $\checkmark$ & \textit{n.s.} \\

fixation\_entropy
& $\checkmark$ & \textit{n.s.} & \textit{n.s.}
& $\checkmark$ & $\checkmark$ & \textit{n.s.}
& \textit{n.s.} & $\checkmark$ & \textit{n.s.} \\

fixation\_scanpath\_length
& -- & -- & --
& -- & -- & --
& \textit{n.s.} & $\checkmark$ & $\checkmark$ \\

saccade\_amplitude\_mean
& -- & -- & --
& $\checkmark$ & \textit{n.s.} & \textit{n.s.}
& \textit{n.s.} & $\checkmark$ & \textit{n.s.} \\

saccade\_count
& $\checkmark$ & \textit{n.s.} & \textit{n.s.}
& $\checkmark$ & $\checkmark$ & \textit{n.s.}
& \textit{n.s.} & $\checkmark$ & $\checkmark$ \\

saccade\_duration\_max
& -- & -- & --
& -- & -- & --
& \textit{n.s.} & $\checkmark$ & \textit{n.s.} \\

saccade\_duration\_percentage
& -- & -- & --
& \textit{n.s.} & $\checkmark$ & \textit{n.s.}
& -- & -- & -- \\

saccade\_duration\_std
& -- & -- & --
& -- & -- & --
& \textit{n.s.} & $\checkmark$ & \textit{n.s.} \\

time\_to\_first\_fixation
& -- & -- & --
& \textit{n.s.} & $\checkmark$ & \textit{n.s.}
& -- & -- & -- \\

total\_fixation\_duration
& $\checkmark$ & \textit{n.s.} & \textit{n.s.}
& -- & -- & --
& \textit{n.s.} & $\checkmark$ & $\checkmark$ \\

\bottomrule
\end{tabular}

\caption{Summary of Tukey post-hoc comparisons for gaze-only features. $\checkmark$ indicates a significant pairwise difference after Tukey adjustment, \textit{n.s.} denotes a tested but non-significant comparison, and -- indicates that the feature was not included in post-hoc analysis for that perception dimension. H--L, H--M, and L--M denote High--Low, High--Medium, and Low--Medium comparisons, respectively.}
\label{tab:gaze_tukey_summary}

\end{table*}

\begin{table*}[t]
\centering
\small
\setlength{\tabcolsep}{6pt}

\begin{tabular}{lccc ccc ccc}
\toprule
& \multicolumn{3}{c}{\textbf{Wealthy}} 
& \multicolumn{3}{c}{\textbf{Safe}} 
& \multicolumn{3}{c}{\textbf{Boring}} \\
\cmidrule(lr){2-4} \cmidrule(lr){5-7} \cmidrule(lr){8-10}
\textbf{AOI}
& \textbf{H--L} & \textbf{H--M} & \textbf{L--M}
& \textbf{H--L} & \textbf{H--M} & \textbf{L--M}
& \textbf{H--L} & \textbf{H--M} & \textbf{L--M} \\
\midrule

road
& $\checkmark$ & $\checkmark$ & $\checkmark$
& $\checkmark$ & \textit{n.s.} & $\checkmark$
& $\checkmark$ & $\checkmark$ & \textit{n.s.} \\

sidewalk
& -- & -- & --
& -- & -- & --
& $\checkmark$ & \textit{n.s.} & \textit{n.s.} \\

building
& $\checkmark$ & \textit{n.s.} & $\checkmark$
& -- & -- & --
& $\checkmark$ & $\checkmark$ & \textit{n.s.} \\

wall
& $\checkmark$ & $\checkmark$ & $\checkmark$
& $\checkmark$ & $\checkmark$ & $\checkmark$
& $\checkmark$ & $\checkmark$ & \textit{n.s.} \\

fence
& $\checkmark$ & $\checkmark$ & $\checkmark$
& $\checkmark$ & $\checkmark$ & $\checkmark$
& $\checkmark$ & \textit{n.s.} & \textit{n.s.} \\

pole
& $\checkmark$ & $\checkmark$ & \textit{n.s.}
& $\checkmark$ & \textit{n.s.} & $\checkmark$
& $\checkmark$ & \textit{n.s.} & \textit{n.s.} \\

traffic light
& \textit{n.s.} & $\checkmark$ & $\checkmark$
& -- & -- & --
& \textit{n.s.} & \textit{n.s.} & $\checkmark$ \\

traffic sign
& \textit{n.s.} & $\checkmark$ & $\checkmark$
& \textit{n.s.} & $\checkmark$ & \textit{n.s.}
& -- & -- & -- \\

vegetation
& $\checkmark$ & $\checkmark$ & $\checkmark$
& $\checkmark$ & $\checkmark$ & $\checkmark$
& $\checkmark$ & $\checkmark$ & \textit{n.s.} \\

terrain
& -- & -- & --
& -- & -- & --
& $\checkmark$ & $\checkmark$ & \textit{n.s.} \\

sky
& $\checkmark$ & $\checkmark$ & $\checkmark$
& $\checkmark$ & $\checkmark$ & $\checkmark$
& $\checkmark$ & $\checkmark$ & $\checkmark$ \\

person
& $\checkmark$ & \textit{n.s.} & \textit{n.s.}
& -- & -- & --
& $\checkmark$ & $\checkmark$ & $\checkmark$ \\

rider
& -- & -- & --
& -- & -- & --
& $\checkmark$ & \textit{n.s.} & \textit{n.s.} \\

car
& $\checkmark$ & \textit{n.s.} & $\checkmark$
& $\checkmark$ & \textit{n.s.} & $\checkmark$
& \textit{n.s.} & \textit{n.s.} & $\checkmark$ \\

truck
& $\checkmark$ & $\checkmark$ & $\checkmark$
& $\checkmark$ & $\checkmark$ & \textit{n.s.}
& $\checkmark$ & $\checkmark$ & \textit{n.s.} \\

bus
& -- & -- & --
& \textit{n.s.} & \textit{n.s.} & \textit{n.s.}
& -- & -- & -- \\

train
& $\checkmark$ & \textit{n.s.} & \textit{n.s.}
& $\checkmark$ & \textit{n.s.} & $\checkmark$
& -- & -- & -- \\

motorcycle
& \textit{n.s.} & \textit{n.s.} & $\checkmark$
& -- & -- & --
& $\checkmark$ & \textit{n.s.} & \textit{n.s.} \\

\bottomrule
\end{tabular}

\caption{Summary of Tukey post-hoc comparisons for AOI gaze time share ($t\_share$). $\checkmark$ indicates a significant pairwise difference after Tukey adjustment, \textit{n.s.} denotes a tested but non-significant comparison, and -- indicates that the AOI was not included in post-hoc analysis for that perception dimension. H--L, H--M, and L--M denote High--Low, High--Medium, and Low--Medium comparisons, respectively.}
\label{tab:aoi_tukey_summary}
\end{table*}

\end{document}